\documentclass[default]{sn-jnl}

\jyear{2022}%

\theoremstyle{thmstyleone}%
\theoremstyle{thmstyletwo}%

\theoremstyle{thmstylethree}%
\newtheorem{definition}{Definition}%

\begin{document}

\title[Automatic tagging of knowledge points for K12 math problems]{Automatic tagging of knowledge points for K12 math problems}


\author[1]{\fnm{Xiaolu} \sur{Wang}}\email{71194501185@stu.ecnu.edu.cn}

\author[1]{\fnm{Ziqi} \sur{Ding}}\email{10181900107@stu.ecnu.edu.cn}

\author[1]{\fnm{Liangyu} \sur{Chen}}\email{lychen@sei.ecnu.edu.cn}

\affil[1]{\centering\orgdiv{Shanghai Key Laboratory of Trustworthy Computing}\\ \orgname{East China Normal University}, \orgaddress{\city{Shanghai}, \country{China}}}


\abstract{Automatic tagging of knowledge points for practice problems is the basis for managing question bases and improving the automation and intelligence of education. Therefore, it is of great practical significance to study the automatic tagging technology for practice problems. However, there are few studies on the automatic tagging of knowledge points for math problems. Math texts have more complex structures and semantics compared with general texts because they contain unique elements such as symbols and formulas. Therefore, it is difficult to meet the accuracy requirement of knowledge point prediction by directly applying the text classification techniques in general domains. In this paper, K12 math problems taken as the research object, the LABS model based on label-semantic attention and multi-label smoothing combining textual features is proposed to improve the automatic tagging of knowledge points for math problems. The model combines the text classification techniques in general domains and the unique features of math texts. The results show that the models using label-semantic attention or multi-label smoothing perform better on precision, recall, and F1-score metrics than the traditional BiLSTM model, while the LABS model using both performs best. It can be seen that label information can guide the neural networks to extract meaningful information from the problem text, which improves the text classification performance of the model. Moreover, multi-label smoothing combining textual features can fully explore the relationship between text and labels, improve the model's prediction ability for new data and improve the model's classification accuracy.}




\maketitle
\section{Introduction}\label{sec1}

In recent years, with the combination of education and information technology, online education has shown a booming development trend, and the number of online practice problems has massively surged. How to efficiently organize and manage these test resources and effectively realize intelligent processes such as question recommendation, creating question papers quickly and adaptive testing is becoming more and more critical in this field. The automatic tagging of knowledge points of practice problems is the basis for managing question bases and improving the automation and intelligence of education. First, automatic tagging of knowledge points can assist or completely replace manual tagging, effectively reducing teachers' workload and improving the efficiency of tagging. Second, automatic tagging can reduce individual bias caused by subjective factors and improve the accuracy of tagging. Therefore, it is of great practical importance to study the automatic tagging of knowledge points.

\begin{figure*}[h]%
	\centering
	\includegraphics[width=1\textwidth]{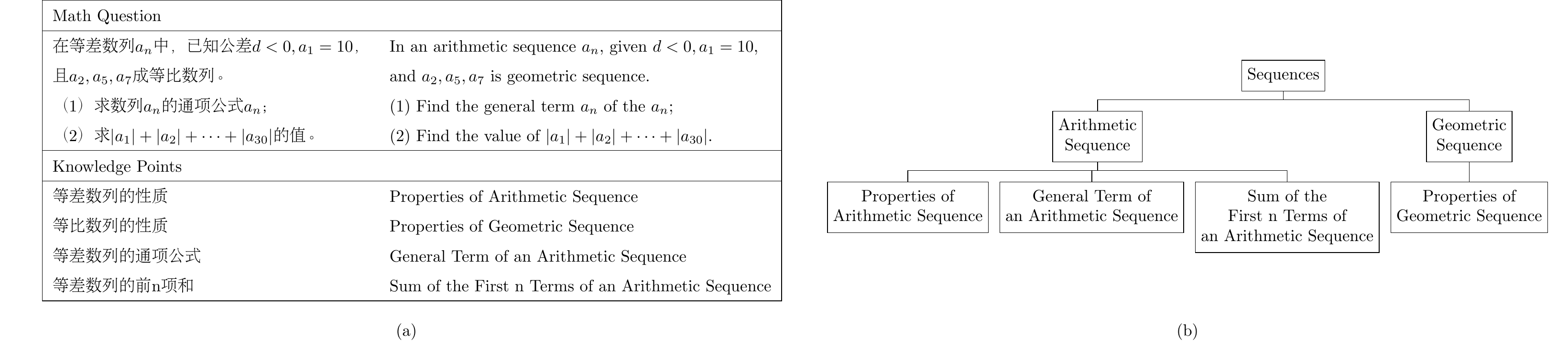}
	\caption{(a) A math question and it's knowledge points (b) A hierarchical relationship of these knowledge points}
	\label{fig:question}
\end{figure*}

As one of the primary subjects in K12 education, mathematics deserves to be researched with emphasis. Math tests are used to test students' mastery of knowledge, and a math problem usually has many different categories of knowledge points. As shown in Figure \ref{fig:question},  these different knowledge points of the question contain a hierarchical relationship. In addition, math texts are usually concise and contain various mathematical symbols with implicit logic and correlations. Moreover, mathematical language is rigorous, and changing one word may describe the opposite result. Traditional text classification methods are difficult to apply directly to math texts to meet the accuracy requirement. Therefore, we need to solve the following issues:

\begin{itemize}
\item Classification of texts containing mathematical formulas
\item Classification of multi-label texts whose labels have correlations
\end{itemize}

In order to do this, the LABS model based on label-semantic attention and multi-label smoothing combining textual features is proposed to improve the automatic tagging of knowledge points for math problems. The main contributions of this paper are as follows:

\begin{itemize}
\item The mathematical objects such as formulas are treated as a whole. The features of the mathematical objects are extracted using the neural network to enrich the semantic features of the text.
\item The novel LABS model is better at predicting the knowledge points of math problems compared with the traditional BiLSTM model.
\item A real-world open source dataset is established for the research. The dataset consists of high school math questions and corresponding knowledge points. The questions contain textual information and mathematical expressions.
\end{itemize}

\section{Related Work}\label{sec2}

The automatic tagging of knowledge points for math problems is essentially the multi-label classification of math texts. The length of a math text is generally within 200 characters, so math texts are short compared to general texts. The key to short text classification is to solve the problem that traditional models cannot extract enough semantic features due to data sparsity. Some studies extend the original text with the help of external knowledge bases\cite{banerjee2007clustering,hu2009exploiting,liu2010short,chen2019deep}, while some directly use the original features of words, such as n-grams\cite{zhang2015short} and word embedding\cite{Meng2017short, Zhang2017research}, to extend the short text. Recently, Li et al.\cite{li2021merging} investigated the effect of fusing statistical information of text with semantic features on short text classification. In addition, Xiao et al.\cite{xiao2019label} used the relationship between labels and texts to build a label-specific document representation, which improved the classification effect of the model. However, these methods are only for texts in general domains and cannot be directly applied to classifying math texts.

There are two main methods in multi-label classification, one is to convert multi-label classification tasks into binary classification tasks or multi-class classification tasks \cite{boutell2004learning,read2011classifier,furnkranz2008multilabel,tsoumakas2007random}, and the other is to deal with multi-label classification problems by extending specific classification algorithms\cite{zhang2007ml,clare2001knowledge,elisseeff2001kernel,mccallum2004collective}. However, there are correlations among labels, and the difficulty of multi-label classification is how to deal with such correlations. Some recent work \cite{yang2018sgm,tsai2019order} converts the multi-label classification into a sequence generation problem using the Seq2Seq model, which uses neural networks to learn the label sequences from text sequences and thus learn the correlation among labels. However, this method requires prior knowledge of the label ordering and has exposure bias, which is not conducive to practical applications. In addition, the label distribution itself implies the relationship among labels, and Guo et al.\cite{guo2020label} let the model learn a simulated label distribution to replace the one-hot representation of labels, which has achieved good effects on multi-class classification. However, there is no relevant study on multi-label text classification. 

\section{LABS Model}\label{sec3}

\subsection{Problem}\label{subsec3.1}

In this paper, the automatic tagging of knowledge points for math problems is summarized as a multi-label classification problem for text with mathematical formulas, which is defined explicitly as 

\begin{definition}[]
$X=\left \{ x_{1},x_{2},\ldots,x_{n} \right \} $ denotes the sequence of a math question, where $ n $ is the length of sequence,$x \in \{x^{w},x^{e}\}$, $x^{w}$ is the word, $x^{e}$ is the mathematical expression. For each input sequence, there is a corresponding knowledge point sequence $Y = \left \{ y_{1},y_{2},\ldots,y_{i},\ldots,y_{l} \right \} $ as an output, where $l$ is the total number of knowledge points, $y_i\in \left \{ 0,1 \right \}$. So the problem is described as $X \rightarrow Y$, i.e., given the question $X$ and its corresponding knowledge points $Y$, the goal is to train a classifier that assigns the most relevant knowledge points for the upcoming new questions.

\end{definition}

\subsection{Solution}\label{subsec3.2}

\begin{figure*}[t]
	\centering
	\includegraphics[width=0.7\textwidth]{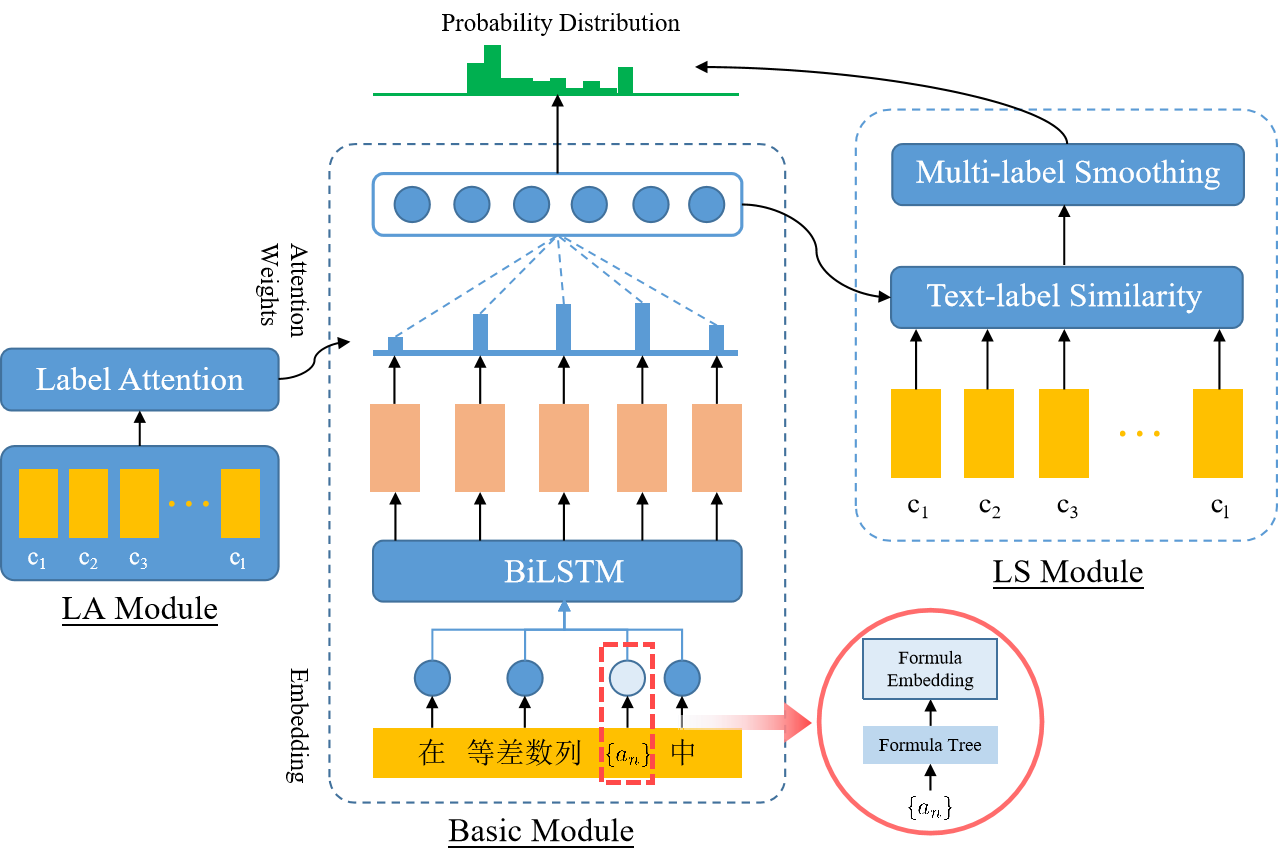}
	\caption{The architecture of the LABS model}
	\label{fig:atmk}
\end{figure*}

To solve the problem above, we propose a model named LABS (Label Attention - Basic - Label Smoothing) in this paper. The architecture of the model is shown in Figure \ref{fig:atmk}.

The LABS model is designed to study the effects of label-semantic attention and multi-label smoothing combining textual features on knowledge points tagging. It consists of three main components:
\begin{itemize}
\item Text representation of math questions (Basic module), described in detail in subsection \ref{subsubsec3.2.1}.
\item Label-semantic attention (LA module), described in detail in subsection \ref{subsubsec3.2.2}.
\item Multi-label smoothing combining textual features (LS module), described in detail in subsection \ref{subsubsec3.2.3}.
\end{itemize}

\subsubsection{Text representation of math questions}\label{subsubsec3.2.1}

The Basic module provides the vector representation for a math question. For example, the math question shown in the Figure \ref{fig:question} includes both general textual content and mathematical objects such as ``$ \{ a_{n} \}$''. If the latter is treated like the former, the mathematical objects will be split like texts, and we will get sequences of words such as ``\{'', ``an'', ``\}''. Such a representation destroys the implicit logic and relationship in mathematical formulas, and no useful information can be extracted. If the information in the mathematical formulas cannot be fully utilized, it is more difficult to achieve effective tagging of the math problems that are brief and concise. For this reason, the LABS model treats mathematical objects such as formulas as a whole, parsing and embedding mathematical formulas based on the TangentCFT method \cite{10.1145/3341981.3344235} and uses neural networks to extract features of mathematical objects for improving the effectiveness of the classifier.

The model uses the BiLSTM network to encode the context of the text from both front and back directions, thus fusing the contextual semantics of both directions into the text representation. For the input sequence $ \left \{ x_{1},x_{2},\ldots,x_{n}  \right \} $ of the math question, its hidden state at time step $t$ is determined by both the input at this time and the hidden state at the previous time step, as shown in Equation (\ref{eq:hidden}):
\begin{equation}
	\label{eq:hidden}
	\begin{split}
		&\overrightarrow{h_{t}}=LSTM( \overrightarrow{h_{t-1}},x_{t} ) ,\\
		&\overleftarrow{h_{t}}=LSTM( \overleftarrow{h_{t-1}},x_{t} ) ,
	\end{split}
\end{equation}
where $x_t$ is the word or formula vector of the input text at time step $t$, $\overrightarrow{h_t},\overleftarrow{h_t}  \in R^k $ are the forward and backward hidden vectors respectively ($k$ is the dimension of the hidden layer).

The final text encoded using BiLSTM is shown in the Equation (\ref{eq:text}), $ H \in R^{n \times 2k} $($n$ is the length of the sequence.)
\begin{equation}
	\label{eq:text}
	\begin{split}
		&H=( \overrightarrow{H}, \overleftarrow{H}) ,\\
		&\overrightarrow{H}=( \overrightarrow{h_{1}}, \overrightarrow{h_{2}}, \ldots,\overrightarrow{h_{n}} ) ,\\
		&\overleftarrow{H}=( \overleftarrow{h_{1}}, \overleftarrow{h_{2}}, \ldots,\overleftarrow{h_{n}} ) .
	\end{split}
\end{equation}

\subsubsection{Label-semantic attention}\label{subsubsec3.2.2}

The LA module is responsible for learning the importance weights of words and formulas in the math text using label-semantic attention. The knowledge points of the problem have specific semantics and correspond to the math text, while conventional attention mechanisms rarely use label information to guide the classification. Therefore, label-semantic attention is proposed, which uses the semantic information of the knowledge points to learn the importance weights of words and formulas in the math text and guide the neural network model to extract meaningful information from the math text, thus enhancing text representation.

The label matrix $C \in R^{l \times k}$ ($l$ is the number of labels) is treated as a trainable matrix, and the weights are dynamically updated in the training stage. In this paper, the similarity between label and text is calculated by the dot product of vectors and then passed into the sigmoid function as shown in Equation (\ref{eq:label_att}).
\begin{equation}
	\label{eq:label_att}
	\begin{split}
		&\overrightarrow{A} = sigmoid(C \overrightarrow{H}^{T} ) ,\\
		&\overleftarrow{A} =sigmoid (C \overleftarrow{H}^{T} )  ,
	\end{split}
\end{equation}
where $\overrightarrow{A} ,\overleftarrow{A} \in R^{l \times n} $ are the similarity between forward text and labels, and backward text and labels, respectively.

Eventually, the text representation enhanced by label-semantic attention as shown in Equation (\ref{eq:att_text}), $M \in R^{l \times 2k}$.
\begin{equation}
	\label{eq:att_text}
	\begin{split}
		&M=( \overrightarrow{M}, \overleftarrow{M}) , \\
		&\overrightarrow{M} =\overrightarrow{A}  \overrightarrow{H}, \\
		&\overleftarrow{M} =\overleftarrow{A}  \overleftarrow{H} .
	\end{split}
\end{equation}

\subsubsection{Multi-label smoothing combining textual features}\label{subsubsec3.2.3}

The LS module is responsible for constructing soft labels and optimizing the classifier using the relationship between the math question and knowledge points. In this paper, the knowledge points of the question are regarded as a trainable matrix. Textual features are integrated into the label distribution using the relationship between math text and knowledge points. Then multi-label smoothing is applied to the classifier, and the loss function is modified to improve the classification effect of the model.

In this paper, the sigmoid activation function is used in the last layer to get the multi-label prediction of the input text. It works as shown in the Equation (\ref{eq:label}): 
\begin{equation}
	\label{eq:label}
	\begin{split}
		&M^{'}=\frac{1}{L}M ,\\
		&y^{(p)}=sigmoid(M^{'}) ,
	\end{split}
\end{equation}
where $M^{'}$ is text vector, and $M^{'} \in R^{2k}$, $y^{(p)} $ is the predicted label distribution.

Usually, to avoid the overfitting and overconfident model caused by the multi-hot encoded label vector, label smoothing (LS) is used as a regularization technique. However, the traditional LS method only adds random noise in each dimension without considering the correlation between labels, so the improvement of the model by the LS method is limited. For this reason, in this paper, based on the LCM (Label Confusion Model) \cite{guo2020label} proposed by Guo et al., we incorporate textual features into the label representation and learn a better label distribution than multi-hot in real time for multi-label smoothing (MLS) to further improve the classification of the model.

MLS treats the label matrix $C \in R^{l \times 2k}$ as a trainable matrix and calculates a confusion distribution $y^{(c)} $ reflecting the similarity between labels according to the Equation (\ref{eq:conf}).
\begin{equation}
	\label{eq:conf}
	y^{(c)} = Softmax( C M^{'}) .
\end{equation}

This confusion distribution is then used to adjust the original multi-hot encoding  representation, which works as shown in the Equation (\ref{eq:prod}):
\begin{equation}
	\label{eq:prod}
	\begin{split}
		y^{(s)}=Softmax(y^{(c)} + \alpha y^{(t)}) ,
	\end{split}
\end{equation}
where $y^{(t)}$ is the multi-hot encoding, which is combined and then normalized with $y^{(c)} $ by the hyperparameter $\alpha$ to obtain the final simulated label distribution $y^{(s)} $.

The Kullback–Leibler divergence is then used to calculate the loss function as shown in Equation (\ref{eq:kl_loss}):

\begin{equation}
	\label{eq:kl_loss}
	loss =  \sum\limits_l^L y_{c}^{(s)} \log \left( \frac{y_{c}^{(s)}}{y_{c}^{(p)}} \right)  .
\end{equation}

\subsubsection{Evaluation metrics}\label{subsubsec3.2.4}

In this paper, Precision@k, Recall@k and F1@k are used to measure the correlation between the predicted value and actual value. Precision@k quantifies the correlation in the first k labels of the predicted results and takes the value of [0,1], and the larger the better.
\begin{equation}
	\label{eq:preck}
	Precision@k= \frac{TP@k}{TP@k+FP@k} .
\end{equation}

The Recall@k gives the proportion of the first k labels that are predicted correctly among the actual correct result and takes the value of [0,1], and the larger the better.
\begin{equation}
	\label{eq:recallk}
	Recall@k= \frac{TP@k}{TP@k+FN@k} .
\end{equation}

F1@k is defined as the harmonic mean of Precision@k and Recall@k and takes the value of [0,1], the larger the better. 
\begin{equation}
	\label{eq:f1k}
	F1@k= \frac{2 \times Precision@k \times Recall@k}{Precision@k + Recall@k} .
\end{equation}

\section{Experimental Setting}\label{sec4}

\subsection{Data}\label{subsec4.1}

Since the problem solved by the LABS model is novel, no suitable public benchmark dataset is available currently. As such, we establish a real-world Chinese dataset for the research, which is currently open source\footnote{https://anonymous.4open.science/r/mathdata-D26B}. This dataset, named the DA-20k, was collected from the online question bases\footnote{http://tiku.zujuan.com/}. It consists of the content of the high school math practice problems including text information and mathematical expressions, and the corresponding knowledge points. The statistical details of DA-20k are shown in Table \ref{tab:total_num}.

\begin{table*}[h]
\begin{center}
\caption{Details of the experimental datasets}
\label{tab:total_num}
\begin{tabular}{@{}llllll@{}}
\toprule
Questions & Labels & Avg. Chars & Avg. Words & Avg. Formulas & Avg. Labels \\ \midrule
22498 & 427 & 68.32      & 47.19      & 6.50          & 1.89        \\ \bottomrule
\end{tabular}
\end{center}
\end{table*}

\subsection{Models and parameters}\label{subsec4.2}

Four models are tested above the dataset DA-20k:
\begin{enumerate}
	\item Basic, including only the Basic module, which is served as a blank control group.
	\item LAB, including Basic module and LA module, which is used to study the role of label-semantic attention on the math text classification.
	\item LBS, including Basic module and LS module, which is used to study the impact of multi-label smoothing combining textual features.
	\item LABS, including Basic module, LA module, and LS module, is used to study the comprehensive effect of the two factors.
\end{enumerate}

The four models are all tested under the same conditions except for the different modules used, and the parameters used in the experiment are shown in Table \ref{tab:task_param}.

\begin{table}[h]
\begin{center}
\caption{Experimental parameters}
\label{tab:task_param}
\begin{tabular}{@{}ll@{}}
\toprule
Parameter                       & Value                      \\ \midrule
Number of tokens                  & 72904                     \\
Number of labels                   & 427                       \\
Max sequence length               & 120                       \\
Vector dimension                   & 300                       \\
Hidden layer size                 & 512                       \\
Batch size                 & 512                       \\
Optimizer                     & Adam\cite{kingma2014adam} \\
Learning rate                     & 0.001                     \\
Hyperparameter($\alpha$)  & 4\cite{guo2020label}                         \\ \bottomrule
\end{tabular}
\end{center}
\end{table}

\section{Experimental Results}\label{sec5}

\subsection{Analysis of overall experiments}\label{subsec5.1}

\begin{figure*}[h]%
\centering
\includegraphics[width=1\textwidth]{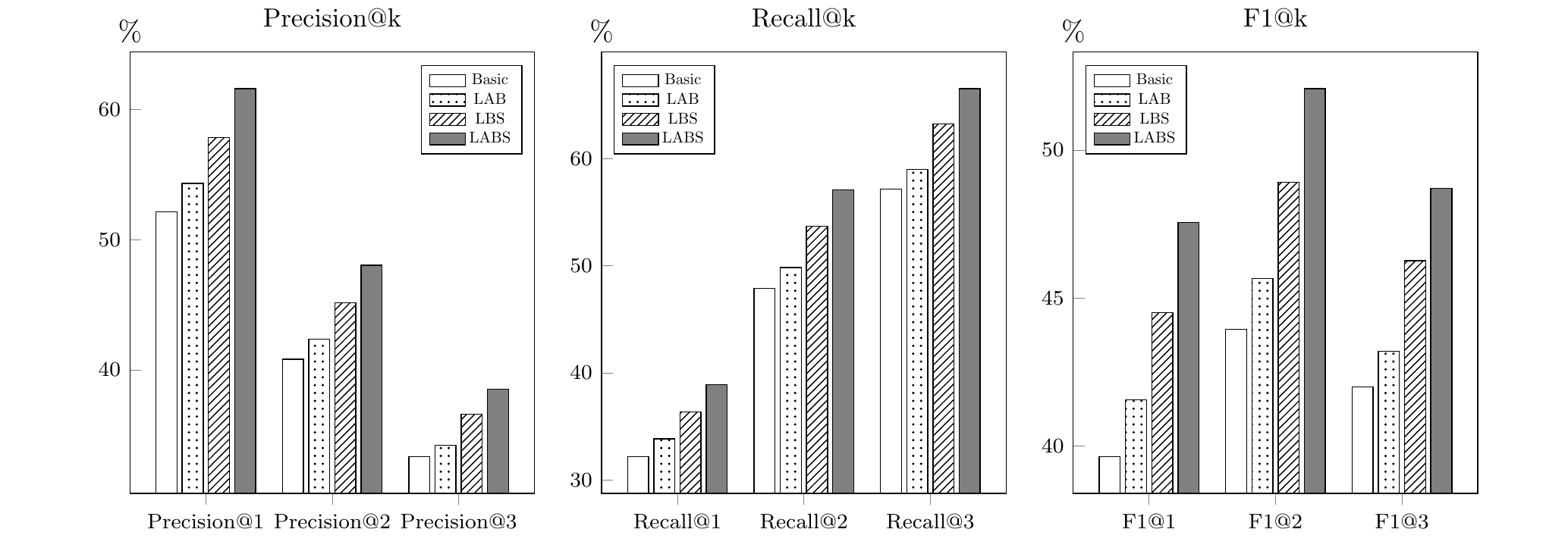}
\caption{Statistical charts of the four models in terms of Precision@k, Recall@k and F1@k (k
= 1,2,3).}\label{fig:result_total}
\end{figure*}

The experimental results of the four models are shown in Table \ref{tab:metric_result} and Figure \ref{fig:result_total}, where the optimal values are rendered in bold.

\begin{table}[h]
\begin{center}
\caption{ Comparison of the four models in terms of Precision@k, Recall@k and F1@k (k = 1,2,3).}
\label{tab:metric_result}
\begin{tabular}{@{}lllll@{}}
\toprule
Evaluation    & Basic   & LAB     & LBS     & LABS             \\ \midrule
Precision@1 & 52.14\% & 54.34\% & 57.84\% & \textbf{61.61\%} \\
Precision@2 & 40.82\% & 42.35\% & 45.14\% & \textbf{48.05\%} \\
Precision@3 & 33.33\% & 34.21\% & 36.58\% & \textbf{38.53\%} \\
Recall@1    & 32.18\% & 33.85\% & 36.36\% & \textbf{38.89\%} \\
Recall@2    & 47.89\% & 49.86\% & 53.70\% & \textbf{57.10\%} \\
Recall@3    & 57.19\% & 59.01\% & 63.26\% & \textbf{66.55\%} \\
F1@1        & 39.65\% & 41.57\% & 44.51\% & \textbf{47.56\%} \\
F1@2        & 43.95\% & 45.67\% & 48.92\% & \textbf{52.07\%} \\
F1@3        & 42.01\% & 43.20\% & 46.26\% & \textbf{48.71\%} \\  \bottomrule
\end{tabular}
\end{center}
\end{table}

In terms of Precision@k, the accuracy sorting of the four models is LABS \textgreater LBS \textgreater LAB \textgreater Basic when the $k$ value stays the same. It is shown that the accuracy of the model prediction is improved after label-semantic attention and multi-label smoothing are introduced. Taking Precision@1 as an example, LAB is improved by 4.22\% over Basic, while LBS is improved by 10.93\%, and the effect of multi-label smoothing is more significant. Furthermore, the LABS model displays the highest performance, improved by 18.16\% over Basic, 13.38\% over LAB, and 6.52\% over LBS. As the $k$ value increases, the accuracy of all four models decreases, indicating that the model has the greatest probability of correctly marking one knowledge point, which is unlikely to exceed the average number of knowledge points of the math practice questions, i.e., 1.89.

In terms of Recall@k, the recall sorting of the four models is LABS \textgreater LBS \textgreater LAB \textgreater Basic when the $k$ value stays the same. It is shown that the model that introduces label-semantic attention or multi-label smoothing has a higher recall under the same case. Taking Recall@3 as an example, LAB is improved by 3.18\% over Basic and LBS is improved by 10.61\%, so the effect of multi-label smoothing is more significant. Furthermore, the LABS model using both has the highest performance, improved by 16.37\% over Basic, 12.78\% over LAB, and 5.20\% over LBS. As the value of $k$ increases, the recall of all four models increases, which indicates that the more knowledge points the model predicts, the more hits the model makes.

In terms of F1@k, the F1-score sorting of the four models is LABS \textgreater LBS \textgreater LAB \textgreater Basic when the $k$ value stays the same. Taking F1@2 as an example, LAB is increased by 3.91\% over Basic and LBS is increased by 11.31\%, so multi-label smoothing was more significant. Furthermore, the LABS model using both has the highest performance, improved by 18.48\% over Basic, 14.01\% over LAB, and 6.44\% over LBS. The F1-score is the harmonic value of precision and recall. The higher this value is, the better the model performs. Thus, LABS is the best model, followed by LBS and LAB, and Basic is the worst. It can be seen that the comprehensive effect of label-semantic attention and multi-label smoothing combining textual features has the greatest impact on the automatic tagging of knowledge points.

\subsection{Distribution of label-semantic attention}\label{subsec5.2}

\begin{figure*}[h]
	\centering
    \includegraphics[width=1\textwidth]{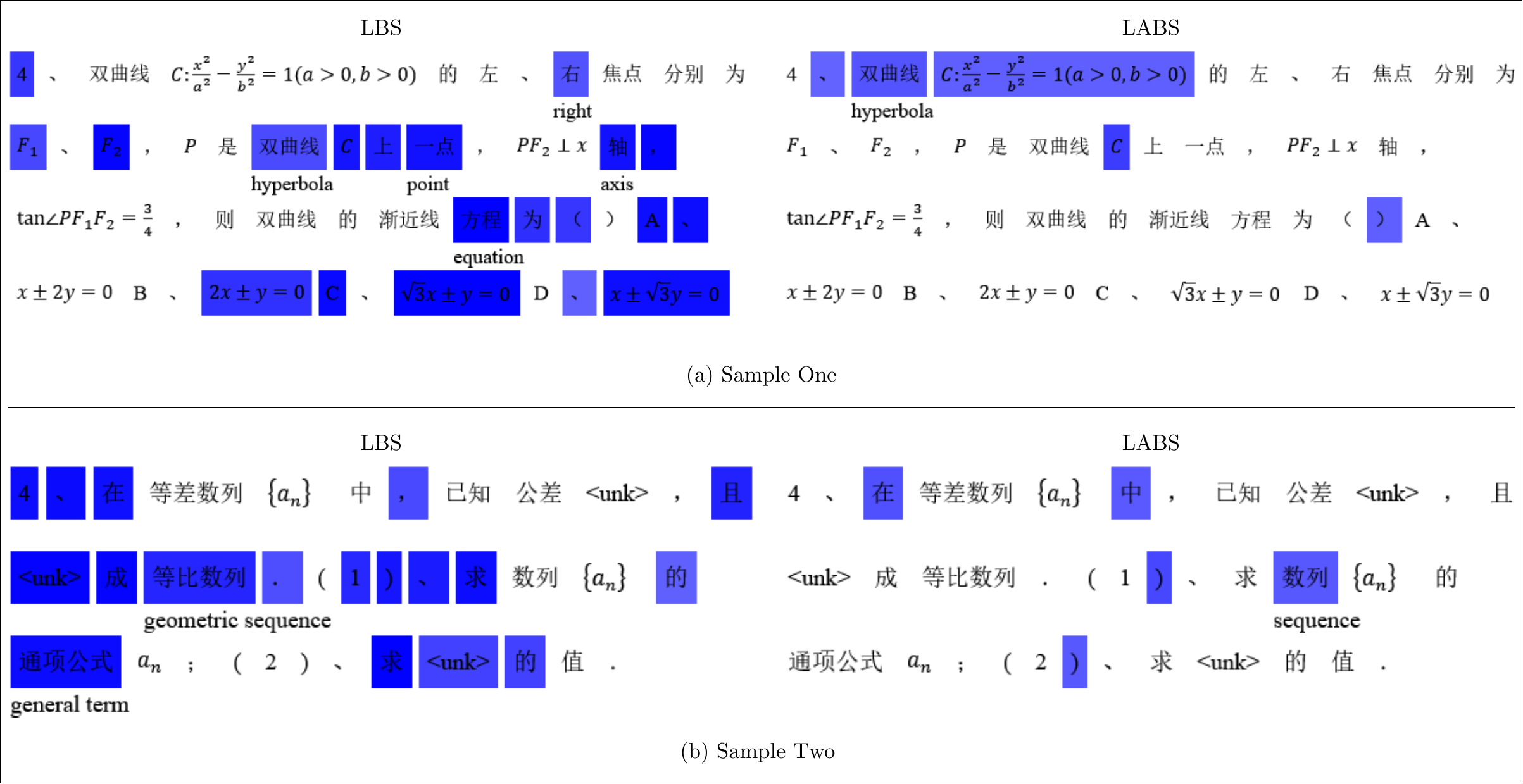}
	\caption{Label attention visualization of math practice problems.  (a) Sample One has labels of \emph{Simple Properties of Hyperbolas} and \emph{Standard Equation for Hyperbola}. (b) Sample Two has labels of \emph{Properties of Arithmetic Sequence}, \emph{Properties of Geometric Sequence}, \emph{General Term of an Arithmetic Sequence} and \emph{Sum of the First n Terms of an Arithmetic Sequence}.}
	\label{fig:attention_dis}
\end{figure*}

In order to better analyze the role of label-semantic attention on the automatic tagging of math problems, we visualize the attention weights on the original math text using heatmaps. The experimental results are shown in Figure \ref{fig:attention_dis} where sample One comes from the dataset, and sample Two is test data shown in Figure \ref{fig:question}. 

As can be seen from the attention distribution, all the knowledge points predicted by the model have related words or formulas given a higher attention weight. For sample One, the LAB model focuses on the words and formulas such as ``hyperbola'' and ``equation'', while the LABS model focuses on ``hyperbola'' and the equation of a hyperbola. For sample Two, the LAB model focuses on ``geometric sequence'' and ``general term'', while the LABS model focuses on ``sequence''. It can be seen that the attention mechanism plays a role. The network model can use the knowledge point semantic information to focus its attention on some part of the math text, which helps improve the model's classification accuracy. Comparing LAB and LABS models, it can be found that the LAB model captures more keywords and the LABS model pays relatively less attention, which indicates that the attention weights are affected by the introduction of multi-label smoothing.

\subsection{Analysis of multi-label smoothing combining textual features}\label{subsec5.3}

\begin{figure*}[h]
\centering
\includegraphics[width=1\textwidth]{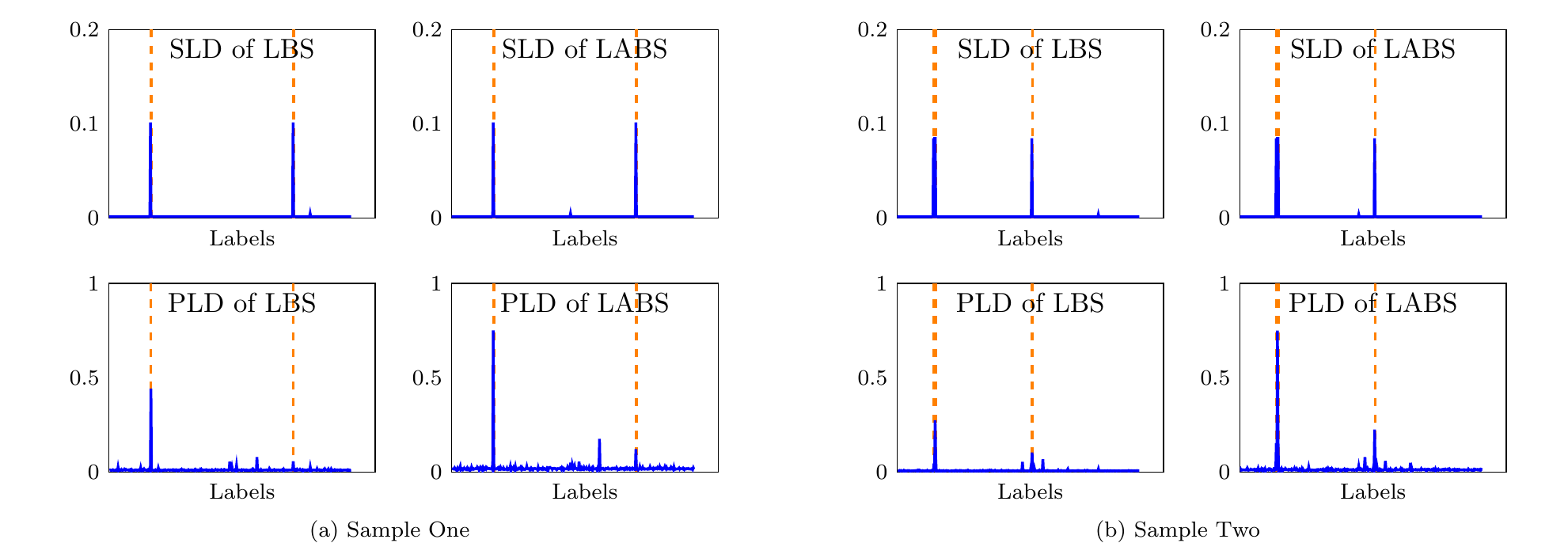}
\caption{The simulated label distribution(SLD) and predicted label distribution(PLD) of two samples on the LBS and LABS model, and the vertical dotted line stands for the actual label distribution.}\label{fig:label_ls}
\end{figure*}

In order to analyze the impact of multi-label smoothing on math text classification, we firstly use the weight of the label simulated layer to calculate the simulated label distribution and then compare it with the actual value and the predicted value. The comparisons are shown in Figure \ref{fig:label_ls}, the sample One and Two are respectively corresponding to samples One and Two in Figure \ref{fig:attention_dis}. 

As can be seen from Figure \ref{fig:label_ls}, compared with the actual label distribution, the value of simulated label distribution is no longer either 0 or 1. The label-text confusion distribution is obtained by similarity measure between text and label, and then it is used to adjust the actual label distribution, resulting in the simulated distribution used for label smoothing. Moreover, Figure \ref{fig:label_ls} also shows that the predicted label distributions of both LBS and LABS are similar to the actual ones, then reflect the strong predictive power of the models. According to the analysis, both LABS and LBS perform better than either LAB or Basic, which indicates that the role of multi-label smoothing is more significant.

\subsection{Comparison of model convergence speed}\label{subsec5.4}

\begin{figure*}[h]
	\centering
	\includegraphics[width=1\textwidth]{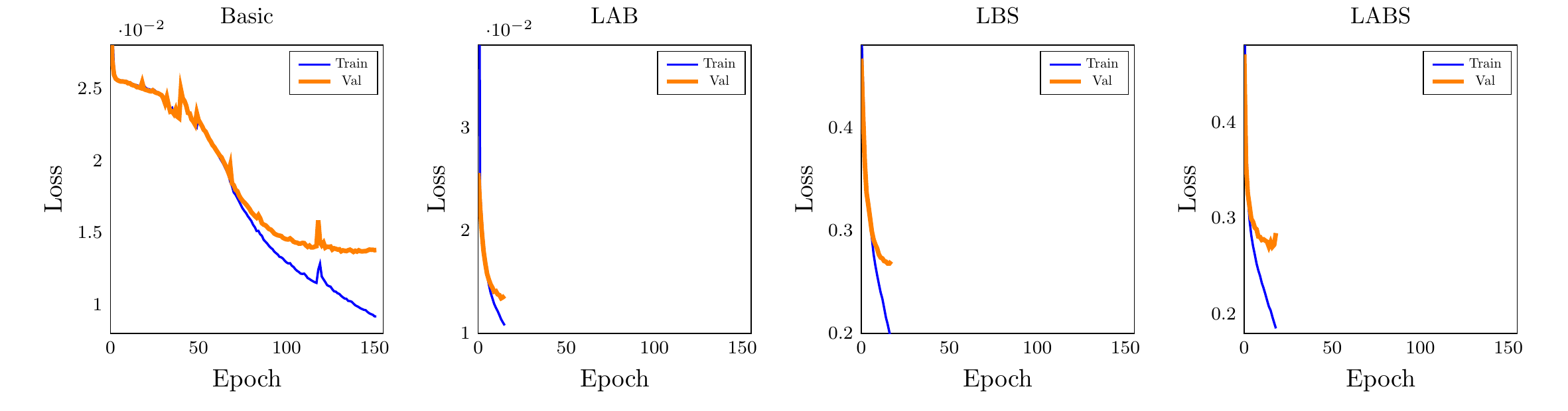}
	\caption{Learning curve of models}
	\label{fig:loss_curve}
\end{figure*}

\begin{figure}[h]
	\centering
	\includegraphics[width=0.35\textwidth]{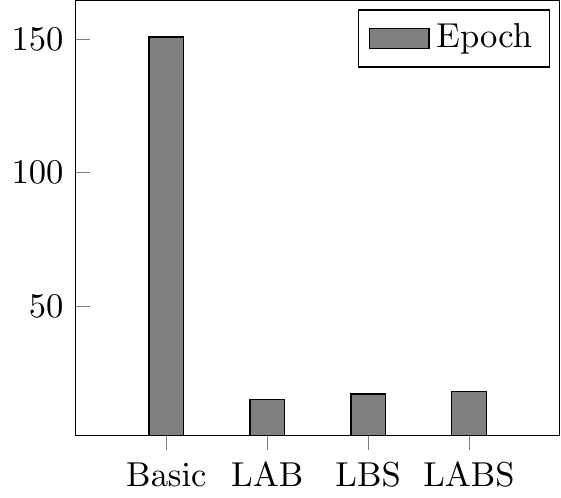}
	\caption{The result of iteration times of models}
	\label{fig:epoch_bar}
\end{figure}

In this paper, the learning curves of the four models are studied, as shown in Figure \ref{fig:loss_curve}. The number of iterations before the early stopping of the four models is shown in Figure \ref{fig:epoch_bar}. 

From the learning curve, we can see that the learning curve of the Basic model is gentle, with low learning efficiency and easy overfitting. However, the learning curve of the other three models is steep, reflecting high learning efficiency, which can effectively avoid overfitting and be easier to obtain models with strong generalization ability. According to the comparisons of iterations, the Basic model converges slowly, and the number of iterations required for training is much higher than that of the other three models. It can be seen that introducing label-semantic attention or multi-label smoothing can accelerate model convergence, improve learning efficiency, and effectively prevent model overfitting.

\subsection{Comparison of different preprocessing methods of mathematical formulas}\label{subsec5.5}

\begin{figure}[h]
	\centering
	\includegraphics[width=0.45\textwidth]{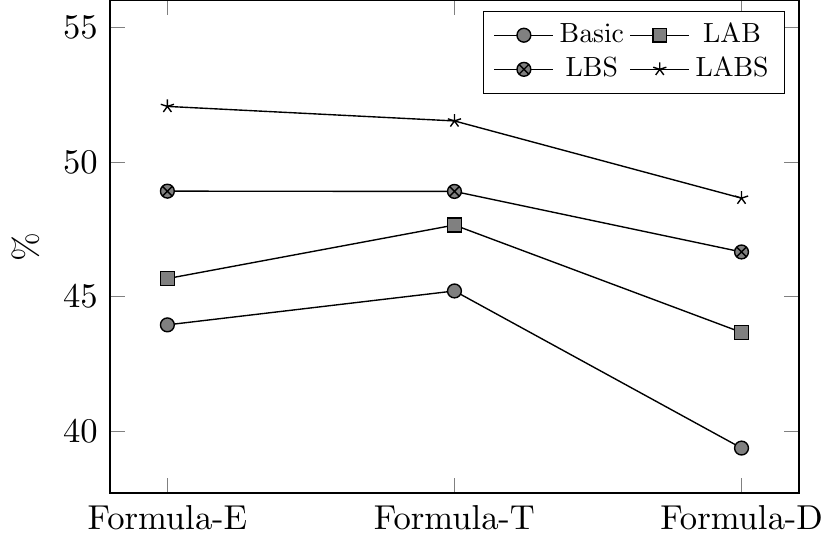}
	\caption{The results of three formula preprocessing methods on four models in terms of F1@2}
	\label{fig:result_formula}
\end{figure}

In addition to formula parsing and embedding (Formula-E), we also consider the other two preprocessing methods: formulas treated as texts (Formula-T) and formula dropping (Formula-D). Then, the effects of these three different treatments were studied on four models. Finally, the model is evaluated on the F1@2 as shown in Figure \ref{fig:result_formula}. 

According to the results, the F1-score of Basic and LAB are all improved by Formula-T compared with Formula-E. Basic is increased by 2.87\%, and LAB is increased by 4.36\%. However, Formula-T has less influence on the LBS and LABS. While with Formula-D, the F1-score decreases substantially in all four models, with 10.42\% in Basic, 4.38\% in LAB, 4.62\% in LBS, and 6.55\% in LABS.

\begin{figure}[h]
	\centering
	\includegraphics[width=0.5\textwidth]{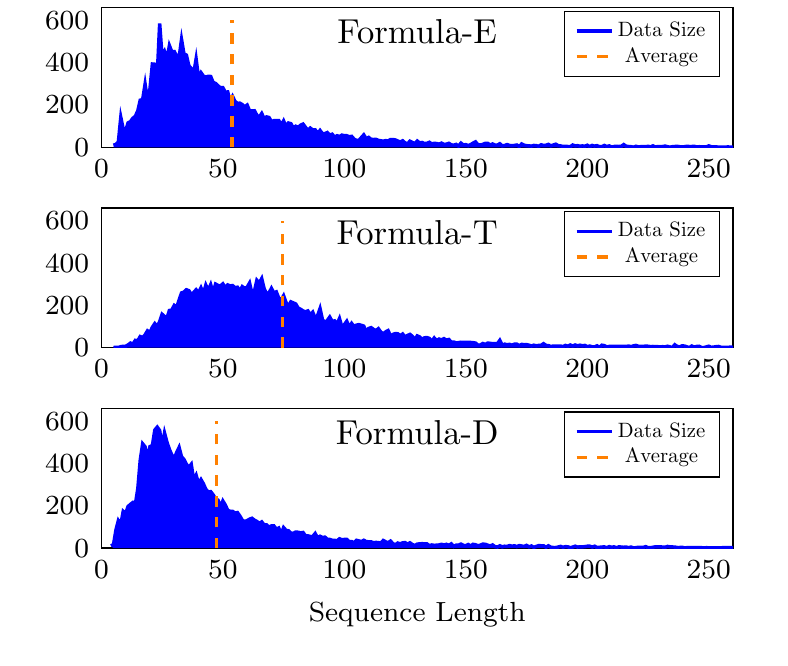}
	\caption{The distribution of input sequence length with three formula preprocessing methods}
	\label{fig:data_dist_result}
\end{figure}

The three preprocessing methods of formulas are intuitively manifested in the average length of the input sequence. As shown in Figure \ref{fig:data_dist_result}, the average length of the input sequence is bigger after Formula-T, while the length distribution after Formula-D and Formula-E are similar. Data detail of math problems is shown in Table \ref{tab:length_result}. It can be seen that Formula-E leads to 6 tokens more than Formula-D, while Formula-T leads to 27 tokens more.

\begin{table}[h]
\begin{center}
\caption{The details of dataset with three formula preprocessing methods}
\label{tab:length_result}
\begin{tabular}{@{}llll@{}}
\toprule
Type & Formula-E & Formula-T & Formula-D \\ \midrule
Avg. words & 47.19    & 74.59    & 47.46  \\
Avg. formulas & 6.50     & 0        & 0      \\
Avg. length & 53.69    & 74.59    & 47.46  \\ \bottomrule
\end{tabular}
\end{center}
\end{table}

Finally, it can be seen that if the formula is ignored and dropped, the classification performance of the model will be worse, so the preprocessing of special elements such as the formula is an indispensable part of the math text representation. In addition, although Formula-T increases the input sequence length and brings some performance increases to the Basic and LAB models, there is no relevant effect on the LABS model proposed in this paper.

\section{Conclusions and Future Work}\label{sec6}

In this paper, we proposed the LABS model based on the BiLSTM model introducing label-semantic attention and multi-label smoothing combining textual features. Experimental results show that the LABS model has optimal precision, recall, and F1-score. However, we do not quantitatively measure the relationship between knowledge points and only conducted experiments on a unique dataset. Besides, the math text has few features and is easy to overfit. In the future, further research will be made on the hierarchical relationship of knowledge points, linking mathematical entities, and expanding other datasets.




\end{document}